\documentclass[letterpaper, 10 pt, conference]{ieeeconf}  % Comment this line out if you need a4paper
\let\labelindent\relax
\usepackage{enumitem}

\usepackage{longtable}
\usepackage{array}
\usepackage{booktabs}
\usepackage[most]{tcolorbox}
\usepackage{multicol}
\usepackage{enumitem}
\usepackage{hyperref}

\IEEEoverridecommandlockouts                              % This command is only needed if 
\title{\LARGE \bf
Accessible and Pedagogically-Grounded Explainability for Human–Robot Interaction: A Framework Based on UDL and Symbolic Interfaces}

\author{Francisco J. Rodríguez Lera$^{1}$, Raquel Fernández Hernández$^{2}$,  Sonia Lopez González$^{2}$,\\ Miguel Angel Gonzalez-Santamarta$^{1}$, Francisco Jesús Rodríguez Sedano$^{1}$, Camino Fernandez Llamas$^{1}$% <-this % stops a space
\thanks{*This work is supported by different national and international organizations}% <-this % stops a space
\thanks{$^{1}$Grupo de Robótica , EIIIA
        Campus de Vegazana, Universidad de León, 24071 León, España
        {\tt\small fjrodl@unileon.es}}%
\thanks{$^{2}$,C.E.E. Ntra. Sra. Del Sagrado Corazón, 
        24005, León, España
        {\tt\small Rfernandezhe@educa.jcyl.es}}%
}

\begin{document}

\maketitle
\thispagestyle{empty}
\pagestyle{empty}

%%%%%%%%%%%%%%%%%%%%%%%%%%%%%%%%%%%%%%%%%%%%%%%%%%%%%%%%%%%%%%%%%%%%%%%%%%%%%%%%
\begin{abstract}

This paper presents a novel framework for accessible and pedagogically-grounded robot explainability, designed to support human–robot interaction (HRI) with users who have diverse cognitive, communicative, or learning needs. We combine principles from Universal Design for Learning (UDL) and Universal Design (UD) with symbolic communication strategies to facilitate the alignment of mental models between humans and robots. Our approach employs Asterics Grid and ARASAAC pictograms as a multimodal, interpretable front-end, integrated with a lightweight HTTP-to-ROS 2 bridge that enables real-time interaction and explanation triggering. We emphasize that explainability is not a one-way function but a bidirectional process, where human understanding and robot transparency must co-evolve. We further argue that in educational or assistive contexts, the role of a human mediator (e.g., a teacher) may be essential to support shared understanding. We validate our framework with examples of multimodal explanation boards and discuss how it can be extended to different scenarios in education, assistive robotics, and inclusive AI.\footnote{Paper under review in RO-MAN 2025 - 01/04/2025}

\end{abstract}

%%%%%%%%%%%%%%%%%%%%%%%%%%%%%%%%%%%%%%%%%%%%%%%%%%%%%%%%%%%%%%%%%%%%%%%%%%%%%%%%
\section{INTRODUCTION}

% Implementar un sistema de explicabilidad para Robótica Social, especialmente dirigido a usuarios con necesidades especiales y utilizando ARASAAC, es una idea muy pertinente y prometedora. Las fuentes proporcionan información valiosa que respalda esta dirección.
The concept of Accessible eXplainable Artificial Intelligence (AXAI) \cite{wolf2020designing} has recently gained attention as a response to the need for more inclusive human–AI interaction. As outlined in \cite{nwokoye2024survey}, AXAI refers to the idea that AI systems should provide clear, understandable explanations of their decisions and behaviors, in ways that are accessible to individuals with varying abilities, backgrounds, and levels of expertise. The aim is to eliminate cognitive and technical barriers, enabling broader segments of the population to engage effectively with AI technologies in their everyday lives.
This perspective aligns with the approach proposed in this paper, in which accessibility and interpretability are integrated through symbolic interfaces and pedagogically grounded design strategies. Rather than treating explainability as a purely technical output, AXAI frameworks emphasize the need to adapt explanations to users' cognitive profiles and communicative preferences.

One of the main explainability issues in social robotics lies in what is known as the perceptual belief attribution problem \cite{Thellman2021}. This problem refers to the difficulty people experience when trying to understand what a robot knows or believes about objects and events in the world, based on its perception\cite{1570532}. Robotic perceptual systems are often not transparent or easily interpretable by non-expert users.

This inability to assess what a robot knows—or does not know—about the shared environment can lead to a range of communicative and interactive issues, such as difficulties in referring to objects or adapting to changes in the environment. The predictability and explainability of robotic systems rely on the human ability to solve the broader intentionality attribution problem, which includes attributing appropriate mental states such as beliefs and desires. The belief attribution problem is a specific subcomponent of this broader challenge.

% \section{Mental Models in Human–Robot Interaction}

A central challenge for researchers and designers in social robotics is to shape human–robot interactions in ways that enable people to understand what robots know about their environment. Achieving this requires a clear understanding of when and why people form incorrect or inadequate mental models of a robot’s perceptual and cognitive mechanisms.

This study investigates how explainable robot behavior can support users with special cognitive or perceptual support needs in forming accurate mental models of a robot’s knowledge and decision-making processes. Our focus is on the role of explanation strategies in aligning the user’s mental model with the robot’s internal belief states, thereby improving mutual understanding, trust, and collaboration.

To address the diverse cognitive and perceptual needs of users in human–robot interaction scenarios, we apply the principles of Universal Design for Learning (UDL)~\cite{hall2003differentiated}to the design of robot explainability strategies. UDL offers a structured framework to ensure that learning and communication processes are accessible to all individuals, particularly those with special support needs. We propose that these principles can be effectively extended to robotic systems, enhancing explainability, transparency, and user engagement.

% The tables below illustrate how each of the three core UDL blocks—representation, action and expression, and motivation—can be mapped to robotic explainability features, along with practical examples.

To support users who may face difficulties with abstract language or conceptual reasoning, it is important to provide explanations that are both accessible and cognitively appropriate. This is where the ARASAAC system (Aragonese Portal of Augmentative and Alternative Communication) can play a key role. Research on ARASAAC pictograms \cite{cabello2015caracteristicas,lopez2017analisis,paolieri2018norms} has shown that they exhibit a high degree of iconicity and are more transparent than other symbol systems, such as SPC or Bliss. This transparency has been replicated across various populations, including adults, typically developing children, and children diagnosed with Autism Spectrum Disorder (ASD).

The importance of high iconicity lies in its association with greater ease of learning and comprehension. In the context of robot explainability, using highly iconic pictograms such as those provided by ARASAAC can significantly facilitate the formation of accurate mental models about the robot’s perceptual beliefs and internal states, especially for users with special support needs.

The main contribution of this work is the design and implementation of an accessible and pedagogically-grounded framework for robot explainability, tailored to users with diverse cognitive and communicative needs. The proposed framework, available in GitHub \url{https://github.com/fjrodl/astericsgrid\_ros},  integrates symbolic multimodal interfaces (based on Asterics Grid and ARASAAC pictograms) with a lightweight bridge to ROS 2 systems, enabling  explainability through real-time interaction. In contrast to existing approaches that focus solely on improving robot transparency, this work emphasizes mutual model alignment and considers the role of human mediators (e.g., teachers or caregivers) within the explanation process. The framework is grounded in principles from Universal Design for Learning (UDL) and validated through practical examples of customizable communication boards that support both user-initiated queries and robot-initiated explanations. This is the first step before deploying robots in different classrooms with different individuals. 

\section{The Perceptual Belief Problem in Human--Robot Interaction}

The \textit{Perceptual Belief Problem} \cite{Thellman2021} refers to the difficulty that humans experience in understanding what a robot perceives and, consequently, what it knows or believes about the shared environment. This issue becomes particularly relevant when users attempt to interpret or anticipate a robot’s decisions or actions based on its internal states.

\begin{figure}[h]
    \centering
    \includegraphics[width=\linewidth]{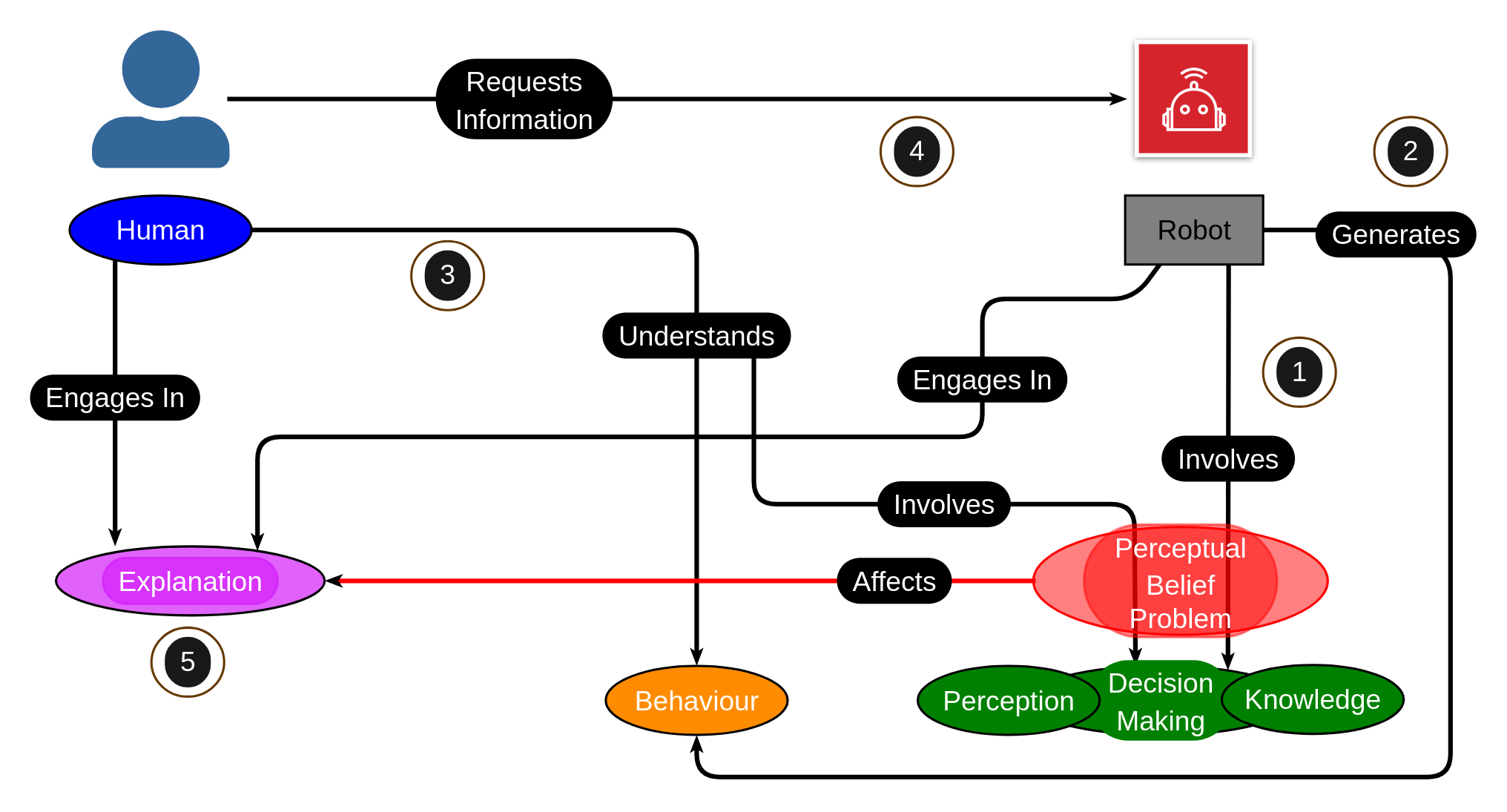}
    \caption{Interaction flow highlighting the Perceptual Belief Problem.}
    \label{fig:belief-problem}
\end{figure}

Figure~\ref{fig:belief-problem} illustrates the core dynamics involved in this problem. The robot engages in perception, decision-making, and knowledge generation (1), which are not always transparent to the human observer. When the robot acts (2), the human attempts to make sense of the robot’s behavior and may explicitly request information (3), prompting the robot to produce explanations (4). These explanations are meant to help the human build an accurate mental model of the robot's beliefs and decision-making process, and ideally, affect their subsequent behavior or trust in the robot (5).

However, misalignments often occur between the robot’s internal representations and the human's assumptions about them. These misalignments are exacerbated in users who may have cognitive, perceptual, or communicative differences—such as those with learning disabilities, language difficulties, or attention-related conditions.

To address these issues, it is essential to design robot explanations that are not only informative, but also accessible and adaptable to the diverse needs of users. In the following section, we propose applying the principles of Universal Design (UD) \cite{mace1997universal} and Universal Design for Learning (UDL) \cite{rao2014review} to robotic explainability systems. These frameworks allow for the creation of explanation strategies that promote perceptual clarity, conceptual accessibility, and inclusive interaction design~\cite{clarkson2013inclusive}.

\section{Universal Design Principles for Inclusive Robotic Explainability}

Universal Design (UD) \cite{mace1997universal} is a design philosophy that aims to make products, environments, and systems usable by all people, to the greatest extent possible, without the need for adaptation or specialized design. Originally developed in the context of architecture and physical accessibility, UD principles have been successfully extended to education, technology, and interaction design. In the context of human–robot interaction, these principles offer a valuable framework for developing explanation mechanisms that are accessible, understandable, and usable by individuals with diverse sensory, cognitive, and communicative profiles.

The core principles and their application to a robotic explainability framework could be enumerated as:

\begin{itemize}
    \item \textbf{Equitable Use:} Explanation strategies should be useful to users with a broad range of abilities. For example, a robot might provide the same explanation content via both spoken language and visual symbols.
    
    \item \textbf{Flexibility in Use:} Systems should support different interaction and explanation preferences, such as allowing users to choose between audio, text, or pictogram-based explanations.
    
    \item \textbf{Simple and Intuitive Use:} Robot explanations should be easy to interpret, regardless of a user’s prior experience or cognitive profile. This includes using plain language and intuitive visual cues.
    
    \item \textbf{Perceptible Information:} The robot must communicate effectively across sensory modalities. For example, critical information should be delivered both visually and auditorily, and designed to stand out in noisy or low-visibility environments.
    
    \item \textbf{Tolerance for Error:} Explanation systems should anticipate and recover from misunderstandings or misinterpretations. A robot might repeat or reformulate its reasoning when detecting user confusion.
    
    \item \textbf{Low Physical Effort:} Explanation delivery should not depend on demanding input methods or prolonged focus. For instance, a robot might adapt the pacing and duration of its explanations based on user engagement.
    
    \item \textbf{Size and Space for Approach and Use:} Interaction zones, displays, and sensors should be physically accessible, and explanations should be readable or hearable from multiple positions.
\end{itemize}

We also draw from inclusive educational design guidelines \cite {SONG2024100212} to enhance the accessibility and usability of robotic explanation systems:

\begin{itemize}
    \item \textbf{Inclusiveness:} Robots should be designed with the expectation that users may have diverse needs. For example, a robot could invite users to customize how explanations are delivered.
    
    \item \textbf{Physical Access:} Robots and their interfaces must be usable by individuals with varying mobility, sensory, or motor profiles.
    
    \item \textbf{Delivery Methods:} Explanation mechanisms should incorporate multiple modalities—spoken, visual, symbolic, tactile—to support diverse preferences and abilities.
    
    \item \textbf{Information Resources:} The robot should offer explanations in a way that allows for previewing, repetition, and review. For instance, a user might ask the robot to repeat an explanation later in the interaction.
    
    \item \textbf{Interaction:} The system should support two-way interaction, allowing users to ask clarification questions, request alternative formats, or provide feedback.
    
    \item \textbf{Feedback:} Robots should provide timely and specific feedback about the user’s understanding or actions, and should be able to confirm whether the explanation was helpful.
    
    \item \textbf{Assessment:} If the robot is being used in educational or therapeutic contexts, it should support various ways to assess user comprehension.
    
    \item \textbf{Accommodation:} Systems should be designed with built-in flexibility to accommodate specific needs, such as visual schedules, simplified modes, or language switching.
\end{itemize}

Applying these principles to robotic explainability contributes to more inclusive and transparent human–robot interactions. By aligning explanation strategies with users’ perceptual and cognitive profiles, robots can support the development of accurate mental models and improve communication, trust, and collaborative outcomes—particularly for individuals with special support needs.

\begin{table*}[h]
\caption{Application of UDL to a Robotic Explainability System}
\label{table_representation}
\begin{center}
\resizebox{\textwidth}{!}{%
\begin{tabular}{|p{0.22\textwidth}||p{0.38\textwidth}|p{0.38\textwidth}|}
\hline
\multicolumn{3}{|c|}{Design Multiple Means of
Engagement}\\ \hline
\hline
\textbf{Subprinciple} & \textbf{Application to Explainability} & \textbf{Practical Example} \\
\hline
1.1 Customization & The robot adapts the level of detail, language, or speed & “I can explain in more detail or in simpler terms. What do you prefer?” \\
\hline
1.2 Auditory Alternatives & Uses text, pictograms, or sign language & In a noisy environment, it displays: “I’m going to point B because the path is clear.” \\
\hline
1.3 Visual Alternatives & Uses voice, sounds, or vibrations & “I’m turning left to avoid an obstacle.” \\
\hline
2.1 Clear Vocabulary & Uses plain language and defines technical terms & “I’m using SLAM: it means I build a map while I move.” \\
\hline
2.4 Languages and Icons & Switches language or uses universal icons & “Language set to English. I am avoiding humans to maintain safety.” \\
\hline
3.1 Connection to Prior Knowledge & Relates decisions to previous experiences & “Remember when you followed me to the warehouse yesterday? I’m doing the same now.” \\ \hline
\hline

\multicolumn{3}{|c|}{Design Multiple Means of Representation}\\ \hline
\hline
4.1 Multiple Inputs & User interacts via voice, touchscreen, gestures, or app & User points to an area; the robot responds: “I will go there because the path is clear.” \\
\hline
5.1 Varied Queries & Questions can be asked via voice, text, or predefined buttons & User presses “Why did you stop?” and the robot replies: “I’m waiting for a person to pass.” \\
\hline
5.3 Adjustable Detail Level & Explanations adapt in complexity to the user’s level & Basic: “I’m turning.” / Expert: “Executing evasive maneuver using the DWB planner.” \\
\hline
6.1 Goal Setting & The robot reports its current goal and allows changes & “My current goal is to take this box to the loading zone. Do you want to change it?” \\
\hline

\multicolumn{3}{|c|}{Design Multiple Means of Action and Expression}\\ \hline
\hline
7.1 User Autonomy & User chooses what and how to receive explanations & “Would you like a summary of my decisions or a step-by-step explanation?” \\
\hline
7.2 Relevance & Connects decisions to user’s goals or interests & “I’m choosing this route so we can get to your destination faster.” \\
\hline
8.2 Tailored Support & Adjusts help based on the user’s familiarity & If the user hesitates: “Would you like me to explain it using images?” \\
\hline
9.3 Feedback & Allows user to rate the quality of the explanation & “Did this explanation help you understand my decision? Press yes or no.” \\
\hline
\end{tabular}
}
\end{center}
\end{table*}

\section{Accessible Explainability Strategies}

Building upon the principles of Universal Design (UD) and Universal Design for Learning (UDL), we developed a set of explainability strategies tailored to diverse user needs. These strategies are organized across three key dimensions: \textit{modality}, \textit{adaptability}, and \textit{interactivity}. Our goal is to support the formation of accurate mental models and facilitate the interpretation of robotic behavior in inclusive, multimodal human–robot interaction.

\textbf{Modality} refers to the use of multiple channels—visual, auditory, textual, or symbolic—to communicate the robot’s internal states and intentions. We emphasize the use of highly iconic pictograms, such as those provided by the ARASAAC system, to enhance semantic transparency. 

\textbf{Adaptability} focuses on the robot’s ability to tailor explanations to the user’s cognitive and perceptual profile. This includes selecting the appropriate level of detail, language complexity, or representation format, and is inspired directly by UDL principles (see Table~\ref{table_representation}).

\textbf{Interactivity} captures the user’s role in actively shaping the explanation process. The robot should allow the user to request clarifications, select preferred modes, and provide feedback on the usefulness of the explanations. These elements contribute to agency and trust.

\begin{figure}[h]
    \centering
    \includegraphics[width=\linewidth]{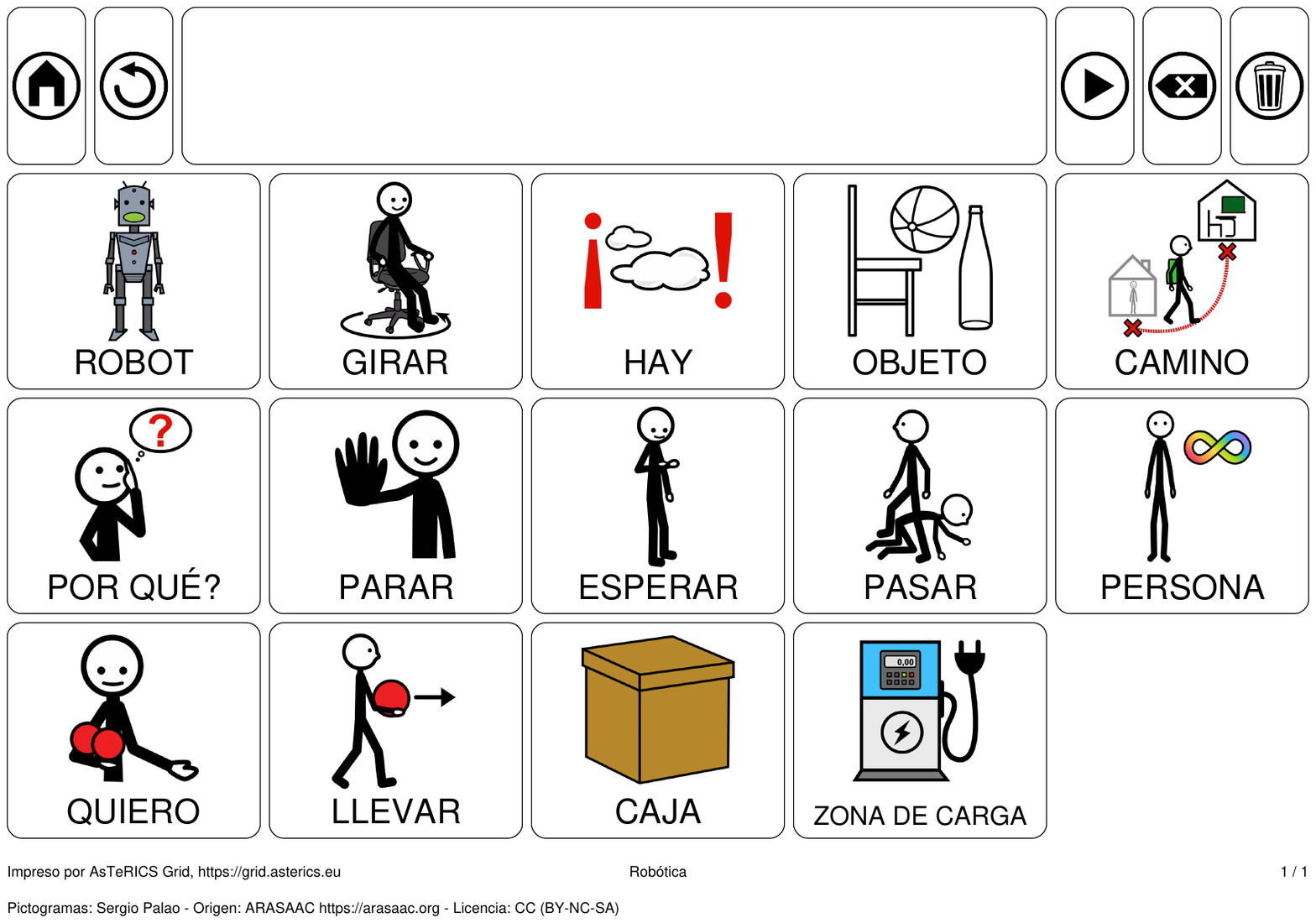}
    \caption{Prototype of a robot explanation interface using ARASAAC pictograms within Asterics Grid.}
    \label{fig:asterics}
\end{figure}

As a practical reference for implementing these strategies, we propose leveraging existing assistive technologies such as \textbf{Asterics Grid}\cite{klaus2024asterics}. This open-source platform enables the creation of customizable communication boards that integrate visual elements (e.g., ARASAAC pictograms), textual labels, and voice output\cite{serrano2024guia}. By embedding robot-generated explanations into a grid-based layout, users can explore the robot’s reasoning in a familiar and accessible format. Figure~\ref{fig:asterics} illustrates an example of this integration.

This combination of inclusive design principles and established assistive technologies provides a flexible and replicable foundation for accessible robotic explainability.

In certain contexts, particularly those involving users with cognitive or communicative support needs, explanation cannot always be achieved through direct interaction between human and robot. Our UDL-informed design anticipates these scenarios by explicitly incorporating the role of a human mediator — such as a teacher, caregiver, or therapist — who can support the explanation loop and facilitate model alignment. From this perspective, explainability becomes not only a functional requirement for interaction but a pedagogical process involving multiple agents. This shift reframes explanation as part of a learning scenario, where the robot, user, and educator collaboratively construct shared understanding.

\section{Framework for Accessible Robotic Explanations}

\begin{figure}[h]
    \centering
    \includegraphics[width=\linewidth]{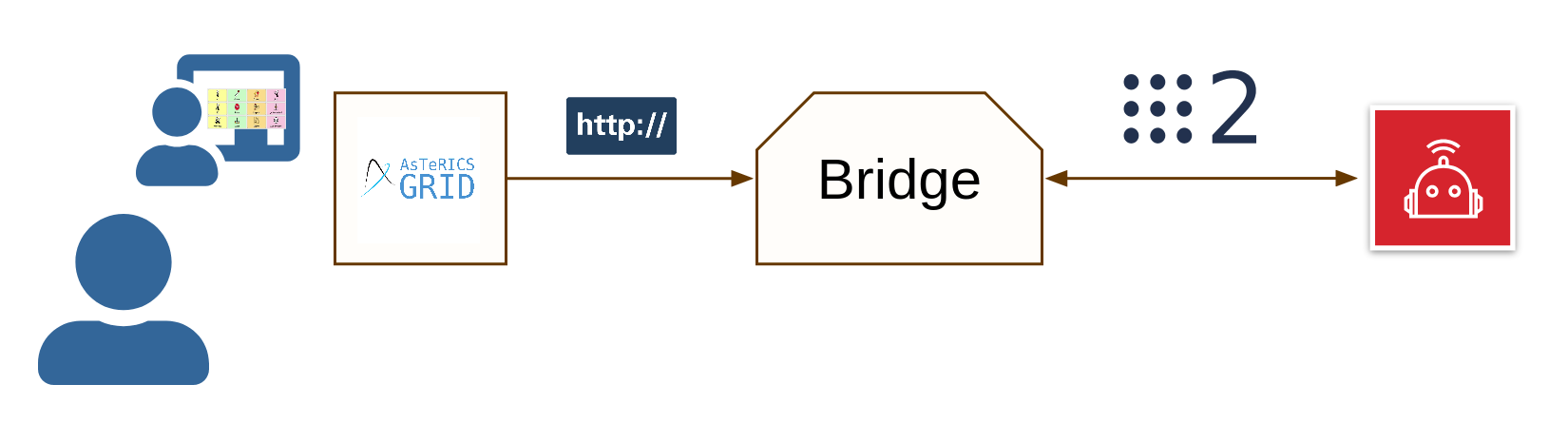}
    \caption{Architecture of the accessible explanation framework connecting Asterics Grid with a ROS 2-based robot.}
    \label{fig:framework}
\end{figure}

To support inclusive and explainable interactions with robotic systems, we developed a modular framework that connects symbolic user interfaces with the robot's cognitive and decision-making layers. The framework follows a low-latency and accessible communication flow structured in five stages (see Figure~\ref{fig:framework}):

\begin{enumerate}
    \item \textbf{Asterics Grid:} A visual board-based interface using ARASAAC pictograms allows users to trigger explanations or ask questions via pictorial and textual elements.
    \item \textbf{HTTP Communication:} Each grid element triggers an HTTP \texttt{POST} request to a local server, sending a payload that identifies the user intention (e.g., \texttt{"turn"}, \texttt{"why"}, \texttt{"goal"}).
    \item \textbf{Flask–ROS 2 Bridge:} A lightweight Python server receives the request, processes the payload, and forwards it as a \texttt{std\_msgs/String} message via ROS 2 to a dedicated topic (\texttt{/asterics\_commands}).
    \item \textbf{ROS 2 Integration:} ROS 2 nodes subscribed to \texttt{/asterics\_commands} interpret the message in context, invoking the corresponding explanatory action or querying the robot’s internal state.
    \item \textbf{Robot Response:} The robot executes the explanation behavior (verbal, visual, gestural) or adjusts its plan, and may respond using the same interface to close the loop.
\end{enumerate}

This design offers several advantages: it is platform-agnostic, modular, and allows for integrating alternative input interfaces or assistive technologies with the standard "de facto" middleware for robots ROS 2~\cite{macenski2022robot}. Furthermore, the use of ARASAAC pictograms ensures semantic transparency and broad accessibility.

While explanation is often conceived as a one-way communication from robot to human, our approach explicitly considers explainability as a bidirectional process. In many cases, especially in educational or assistive contexts, it is not enough for the robot to explain itself — it must also be able to interpret the user's mental model, expectations, or misunderstandings. Effective explanation thus requires mutual model alignment: the user must understand the robot’s perceptual and decision-making processes, while the robot (or the system) must access cues or scaffolding that reveal how the user interprets the situation.

This bidirectionality motivates our decision to include symbolic interfaces that are interpretable by both human and machine, and to allow user-initiated queries that signal information gaps or misunderstandings. Our framework supports this alignment loop as a foundational requirement for human–robot understanding. 

\section{Implementation of the Accessible Explanation Interface}

To support inclusive and explainable human–robot interaction, we developed an explanation interface based on Asterics Grid, a customizable assistive communication platform. The interface is designed to translate the robot’s internal states, intentions, and actions into visual and symbolic explanations using ARASAAC pictograms.

The explanation interface is structured as a dynamic communication board (Figure~\ref{fig:asterics}), where each pictogram represents a key concept involved in the robot’s behavior: actions (e.g., “carry”, “turn”, “wait”), goals (e.g., “charging zone”), agents (e.g., “robot”, “person”), contextual elements (e.g., “object”, “obstacle”), and interaction cues (e.g., “why?”, “stop”, “go”). The board is designed to allow both robot-initiated and user-initiated communication.

% \begin{figure}[h]
%     \centering
%     \includegraphics[width=\linewidth]{figuras/Asterics1.png}
%     \caption{Accessible explanation board using ARASAAC pictograms in Asterics Grid.}
%     \label{fig:grid}
% \end{figure}

The system integrates with the robot’s internal reasoning architecture, enabling real-time explanation generation. The explanation pipeline consists of three layers:

\begin{enumerate}
    \item \textbf{Semantic Mapping:} The robot’s internal events (e.g., navigation goals, planner outputs, sensor triggers) are mapped to a semantic layer containing predefined concepts (e.g., “obstacle ahead”, “carrying object”, “goal changed”).
    
    \item \textbf{Message Construction:} These concepts are transformed into short textual descriptions, which are then matched to corresponding ARASAAC pictograms.
    
    \item \textbf{Grid Presentation:} The selected pictograms are displayed on the grid as sequential messages or as elements the user can query or interact with.
\end{enumerate}

The interface supports both \textit{robot-initiated explanations} (e.g., when the robot changes plan or encounters a problem) and \textit{user-initiated queries} (e.g., the user presses “why?” to trigger a context-aware explanation).

This approach allows us to decouple low-level robot actions from high-level user-relevant representations and to present explanations in a perceptually accessible and cognitively supportive format. It also enables future extensions such as multilingual boards, context-aware pictogram selection, or integration with speech synthesis.

To support inclusive and explainable human–robot interaction, we developed an explanation interface based on Asterics Grid, a customizable assistive communication platform. The interface is designed to translate the robot’s internal states, intentions, and actions into visual and symbolic explanations using ARASAAC pictograms.

The explanation interface is structured as a dynamic communication board, where each pictogram represents a key concept involved in the robot’s behavior: actions (e.g., “carry”, “turn”, “wait”), goals (e.g., “charging zone”), agents (e.g., “robot”, “person”), contextual elements (e.g., “object”, “obstacle”), and interaction cues (e.g., “why?”, “stop”, “go”). The board allows both robot-initiated and user-initiated communication, and can adapt to task context or user profile.

\begin{figure}[h]
    \centering
    \includegraphics[width=\linewidth]{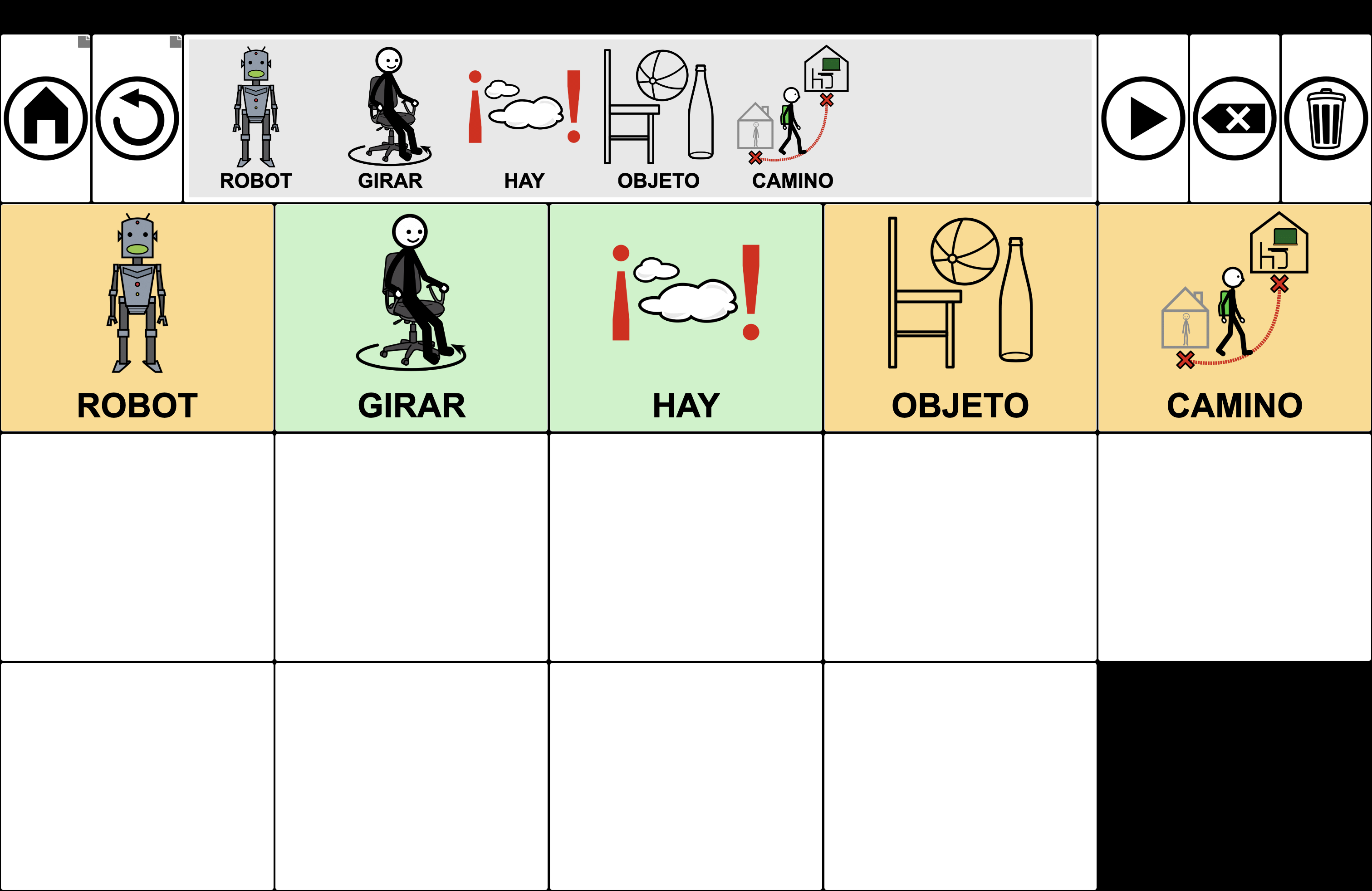}
    \caption{Example 1: User-oriented explanation support and interaction prompts.}
    \label{fig:asterics1}
\end{figure}

\begin{figure}[h]
    \centering
    \includegraphics[width=\linewidth]{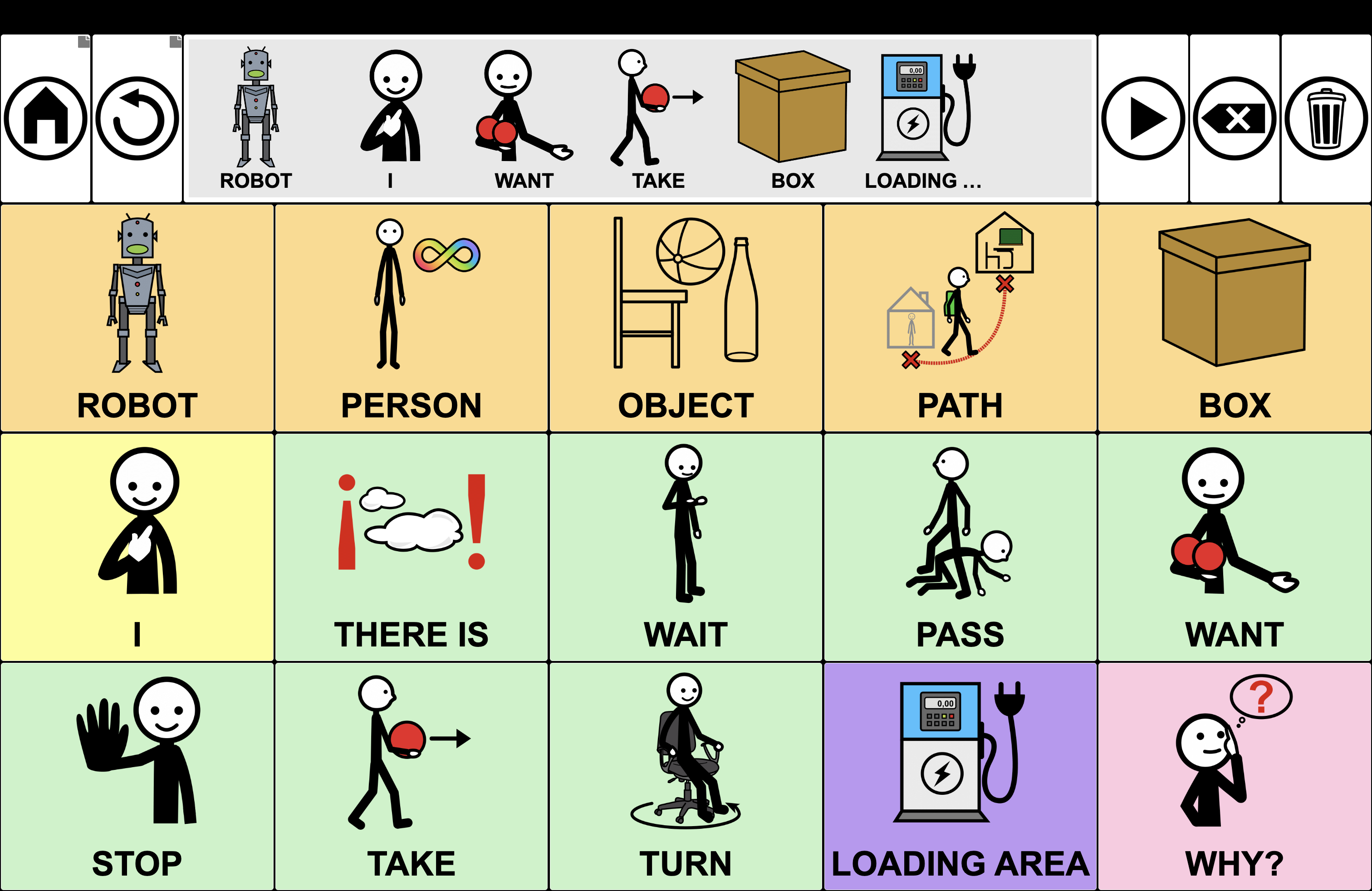}
    \caption{Example 2: Robot-initiated explanation for obstacle avoidance in Spanish.}
    \label{fig:asterics2}
\end{figure}

\begin{figure}[h]
    \centering
    \includegraphics[width=\linewidth]{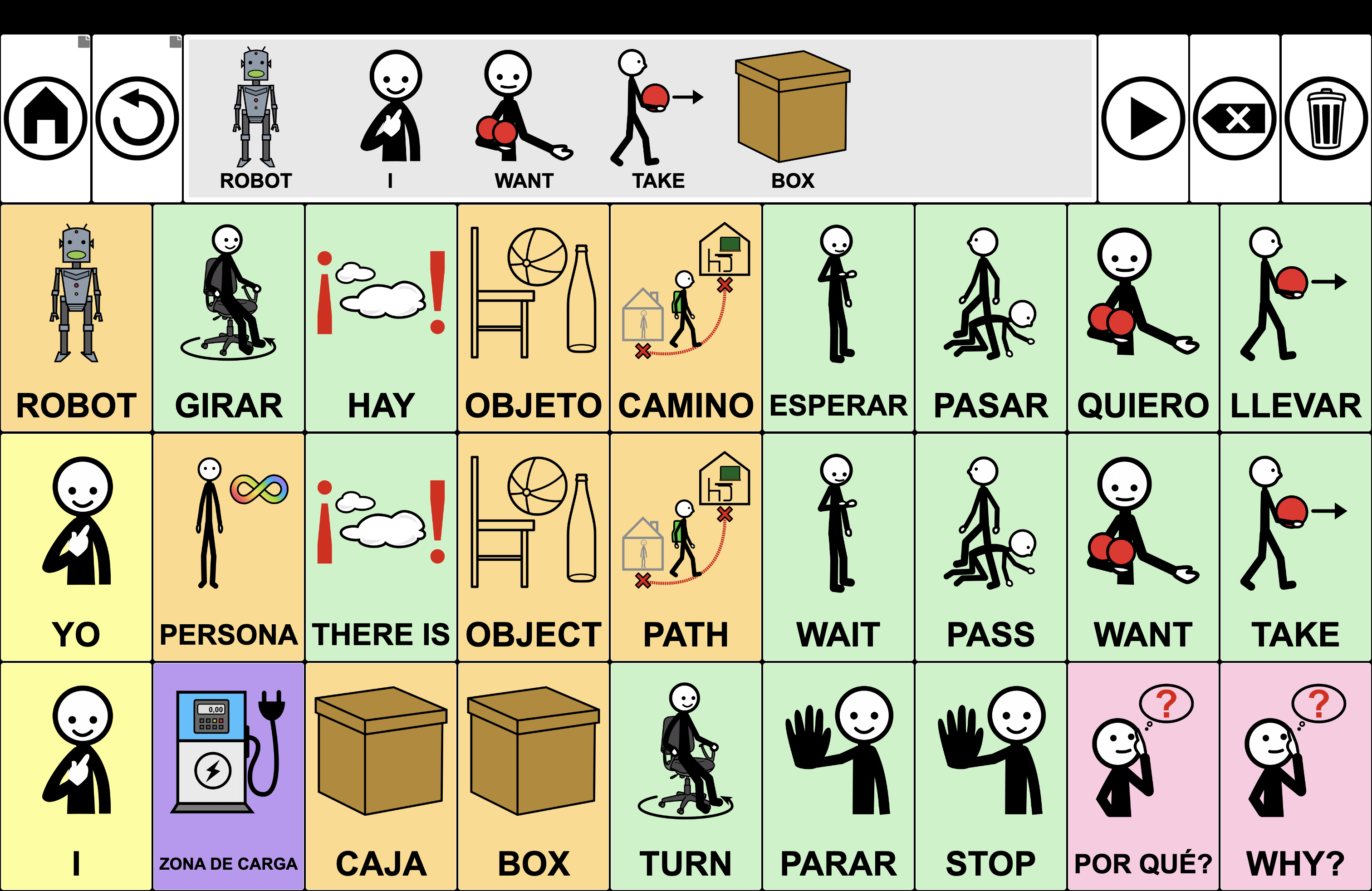}
    \caption{Example 3: Full communication board combining goals, actions, and objects in multiple languages.}
    \label{fig:asterics3}
\end{figure}

Figures~\ref{fig:asterics1}–\ref{fig:asterics3} illustrate three example configurations of this system:

\begin{itemize}
   
    \item \textbf{Figure~\ref{fig:asterics1}:} A focused explanation board displaying the robot’s self-generated message: “Robot turns. There is an object blocking the path.”
    \item \textbf{Figure~\ref{fig:asterics2}:} A simplified interaction panel, supporting quick user queries like “Why?”, “Stop”, “Wait”, or identifying agents and objects.
     \item \textbf{Figure~\ref{fig:asterics3}:} A full grid configuration with action, object, and intent elements, allowing the user to express or understand commands such as “I want to take the box.”
\end{itemize}

These examples demonstrate how accessible explanation strategies can be instantiated in real-time, multimodal interfaces. The visual symbols reduce cognitive load and improve the alignment between the robot’s internal states and the user’s mental model, especially for individuals with communication or processing difficulties.

\section{CONCLUSIONS}

This work proposes an inclusive and modular framework for robot explainability, integrating symbolic interfaces, accessible design principles, and pedagogical scaffolding. Grounded in the principles of Universal Design for Learning (UDL), the framework addresses the need for transparent, adaptable, and user-centered explanation mechanisms in human–robot interaction.

Explainability is treated as a shared cognitive process, in which the robot's internal state must be made interpretable while also adapting to the user's mental model and communicative abilities. In many cases, this process may require the involvement of a human mediator to support mutual understanding and facilitate interaction. The combination of Asterics Grid, ARASAAC pictograms, and ROS 2 integration enables a robust, real-time interface capable of supporting diverse users in both autonomous and collaborative explanation scenarios.

Future developments include deployments in educational and assistive contexts, and the integration of adaptive models to personalize explanations based on user profiles and interaction history.

\addtolength{\textheight}{-12cm}   % This command serves to balance the column lengths

%%%%%%%%%%%%%%%%%%%%%%%%%%%%%%%%%%%%%%%%%%%%%%%%%%%%%%%%%%%%%%%%%%%%%%%%%%%%%%%%



\section*{ACKNOWLEDGMENT}

This work is partially funded under Grant DMARCE (EDMAR+CASCAR) Project PID2021-126592OB-C21 + PID2021-126592OB-C22 funded by MCIN/AEI/10.13039/501100011033 and by ERDF A way of making Europe. This work is partially supported by the Erasmus+ Project ROBOSTEAMSEN - Training SEN teachers to use robotics for fostering STEAM and develop computational thinking with reference: 2023-I-ESOI-KA220-SCH-OOOI55379.

% The authors would like to acknowledge the use of OpenAI's ChatGPT for assistance in the preparation and editing of this manuscript.
The authors gratefully acknowledge the use of OpenAI's ChatGPT for the linguistic review and structural refinement of this manuscript, primarily for the abstract, conclusion, and framework description.

%%%%%%%%%%%%%%%%%%%%%%%%%%%%%%%%%%%%%%%%%%%%%%%%%%%%%%%%%%%%%%%%%%%%%%%%%%%%%%%%

%%% inclusión de referencias
\bibliographystyle{plain}
 \bibliography{roman2025}

\end{document}